\documentclass[letterpaper]{article}
\usepackage[preprint]{aaai2027}
\usepackage[hyphens]{url}
\usepackage{graphicx}
\usepackage{natbib}
\usepackage{caption}
\usepackage{algorithm}
\usepackage{algorithmic}

\usepackage{booktabs}

\usepackage{amsmath}

\usepackage{amssymb}

\usepackage{amsthm}

\newtheorem{theorem}{Theorem}
\newtheorem{proposition}{Proposition}

\newtheorem{corollary}{Corollary}
\theoremstyle{definition}
\newtheorem{definition}{Definition}

\newtheorem{assumption}{Assumption}

\let\Pr\relax
\let\SCOPEsumop\sum\renewcommand{\sum}{\SCOPEsumop\limits}
\let\SCOPEbigcupop\bigcup\renewcommand{\bigcup}{\SCOPEbigcupop\limits}
\let\SCOPEmaxop\max\renewcommand{\max}{\SCOPEmaxop\limits}
\let\SCOPEminop\min\renewcommand{\min}{\SCOPEminop\limits}
\DeclareMathOperator*{\argmax}{arg\,max}
\DeclareMathOperator{\rank}{rank}
\DeclareMathOperator{\spanop}{span}
\DeclareMathOperator{\range}{range}
\DeclareMathOperator*{\Pr}{Pr}
\DeclareMathOperator{\RC}{RC}
\DeclareMathOperator{\Leak}{Leak}
\DeclareMathOperator{\MI}{I}
\newcommand{\E}{\mathop{\mathrm{E}}\limits}

\newcommand{\R}{\mathbb{R}}
\newcommand{\calF}{\mathcal{F}}
\newcommand{\calR}{\mathcal{R}}
\newcommand{\cQ}{\mathcal{Q}}
\newcommand{\UF}{\mathcal{U}_{\calF}}
\newcommand{\hUF}{\hat{\mathcal{U}}_{\calF}}
\newcommand{\SR}{\mathcal{S}_{\calR}}
\newcommand{\Proj}{\mathbf{P}}
\newcommand{\Q}{\mathbf{Q}}
\newcommand{\I}{\mathbf{I}}
\newcommand{\W}{\mathbf{W}}
\newcommand{\Tmat}{\mathbf{T}}
\newcommand{\Hmat}{\mathbf{H}}
\newcommand{\hBF}{\hat{\mathbf{B}}_F}
\newcommand{\Sig}{\boldsymbol{\Sigma}}
\newcommand{\vx}{\boldsymbol{x}}
\newcommand{\vz}{\boldsymbol{z}}
\newcommand{\vw}{\boldsymbol{w}}
\newcommand{\vmu}{\boldsymbol{\mu}}
\providecommand{\vv}{}\renewcommand{\vv}{\boldsymbol{v}}
\newcommand{\vu}{\boldsymbol{u}}
\newcommand{\vb}{\boldsymbol{b}}
\newcommand{\vzero}{\boldsymbol{0}}
\newcommand{\Zero}{\mathbf{0}}
\newcommand{\norm}[1]{\left\lVert #1\right\rVert}
\newcommand{\PUF}{\Proj_{\UF}}
\newcommand{\PcQ}{\Proj_{\cQ}}

\usepackage{multirow}

\DeclareMathSizes{9}{9}{7}{5}
\newcommand{\tablefontsize}{\fontsize{9pt}{10pt}\selectfont}

\newcommand{\tabpm}[2]{{\tablefontsize $#1{\scriptstyle\,\pm\,#2}$}}

\newcommand{\bestpm}[2]{{\tablefontsize $\mathbf{#1}{\scriptstyle\,\boldsymbol{\pm}\,\mathbf{#2}}$}}

\newcommand{\secondpm}[2]{{\tablefontsize $\underline{#1{\scriptstyle\,\pm\,#2}}$}}

\title{SCOPE: Entanglement Frontier Escape for Source-Free Class Unlearning}
\author{
    Junhao~Cai\textsuperscript{\rm 1},
    Dohun~Kim\textsuperscript{\rm 1},
    Sung~Il~Choi\textsuperscript{\rm 1},
    Juhyun~Park\textsuperscript{\rm 1},\\
    Chengjun~Jin\textsuperscript{\rm 1},
    Dowon~Kim\textsuperscript{\rm 1},
    Changhee~Joo\textsuperscript{\rm 1}\corresponding
}
\affiliations{
    \textsuperscript{\rm 1}Korea University, Seoul, Republic of Korea\\
    \{junhochae, dohunkim, sungchoi, juhyunpark, chengjunjin2001, dowonkim, changhee\}@korea.ac.kr
}

\begin{document}

\maketitle

\begin{abstract}
Source-free class unlearning erases whole classes using only the forget data, judged at the representation level, where features can leak a class the head no longer predicts. Existing feature-space erasers answer with one fixed projection, yet forget and retain classes share a representation, so deleting one disturbs the other where they overlap. We prove this tension is a frontier. Every fixed projection that deletes pays a retain cost of at least the retain-readout energy along the forget-discriminant subspace, and erasing that subspace alone attains the floor. The leading source-free erasers all instantiate the form it binds, so the frontier limits the whole class. Conditioning the erasure on the input escapes it. \textbf{S}pectral \textbf{Co}nditional \textbf{P}rojective \textbf{E}rasure (SCOPE) does so with a single gate, suppressing the forget subspace chiefly on inputs its frozen head's weight scores read as a forget class. It is closed form, needs no retain data or gradient training, and costs orders of magnitude less than retraining. Across five object, face, and speaker benchmarks spanning two modalities and both convolutional and transformer backbones, the frontier predicts the measured retain cost. SCOPE leads the source-free erasers on every benchmark and forget-set size, and at the hardest setting it tops every unlearner, trained methods included.
\end{abstract}

\section{Introduction}
\label{sec:intro}

Machine unlearning removes the influence of a designated forget set from a trained model while preserving its behavior on the remaining retain set~\citep{bourtoule2021machine}. Retraining from scratch on the retain set is the most reliable way to achieve this, but is computationally prohibitive at modern scale and, more restrictively, presumes continued access to the original training data~\citep{izzo2021approximate}. These constraints have motivated source-free class unlearning, where entire classes must be erased using only forget data and the trained model, without the retain set or a from-scratch retrain~\citep{esc,delete}. An update that suppresses the forget classes at the output can leave them fully recoverable in feature space, a privacy leak in its own right. The community therefore audits deletion not by test accuracy but by knowledge-retention (KR) re-extraction~\citep{esc}. A fresh linear probe, independent of the deployed head, is retrained on the unlearned features, and the forget classes must no longer be linearly decodable. Passing this re-extraction test is what we mean by deleting a class at the representation level. Section~\ref{sec:theory} formalizes the within-forget core of this notion, that the forgotten classes collapse into one another, while the experiments audit both it and the stronger full-class re-extraction throughout.

\begin{figure}[t]
\centering
\includegraphics[width=\columnwidth]{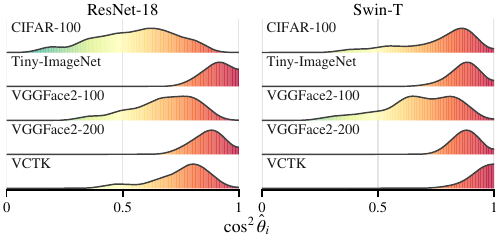}
\caption{Fixed-projection deletion must disturb retention where forget and retain readouts overlap. Each ridge shows the overlaps $\cos^{2}\hat\theta_{i}$, with $\hat\theta_{i}$ the empirical deployed-coordinate principal angles between the forget-discriminant directions ($\hUF$) and the retain-readout subspace ($\SR$) of the original model, formalized in Sections~\ref{sec:theory} and~\ref{sec:scope}, pooled over forget-set sizes $K\in\{2,5,10,20\}$, where $K$ is the number of forgotten classes, with $1$ a direction shared with the retain readout and $0$ one deletable at no retain cost.}
\label{fig:entangle}
\end{figure}

Deletion is hard because the forget and retain classes share a representation. The directions that linearly separate the forget classes can coincide with the directions the retained classifier reads from. How much these two sets overlap, formalized in Section~\ref{sec:theory} through the principal angles between them, weighted by the retained readout energy, sets the deletion cost. When they are orthogonal a class can be removed at no cost to retention, but where they share directions any fixed projection that collapses the forget classes must also disturb the retained readout. Figure~\ref{fig:entangle} shows this overlap holds in every benchmark and backbone. The directions that separate the forget classes lie almost entirely inside the subspace the retained classifier reads from.

Some approaches update the model's parameters~\citep{badt,delete}, but often leave the feature representation close to the original~\citep{kim2025truly}. The training-free methods we study instead edit the representation, projecting the penultimate feature off a fixed subspace to collapse the forget classes, a closed-form global projection that aims to pass the audit with no retain data~\citep{esc,pour}. A global projection applies one fixed map to every input, including the retain inputs it must preserve, so where the forget-discriminant and retain-readout directions overlap it cannot avoid the entanglement cost. This raises a question. \emph{Is the entanglement cost fundamental, or an artifact of the fixed map?}

The cost is fundamental to fixed linear maps but not to the problem itself. We prove an entanglement frontier that binds every input-independent deleting projection (Theorem~\ref{thm:frontier}), and show that conditioning the erasure on the input escapes it (Proposition~\ref{prop:escape}), building retain-neutrality into the conditioning rather than correcting it after a fixed projection. We instantiate this as \textbf{SCOPE (Spectral Conditional Projective Erasure)}, a closed-form source-free class unlearner whose gate, read from the frozen head's weight scores, removes the forget-discriminant directions mainly on inputs read as a forget class, with no retain data or gradient training.

The contributions of our work are as follows:
\begin{itemize}
    \item We prove an entanglement frontier (Theorem~\ref{thm:frontier}), a tight lower bound on the retain cost of every input-independent deleting linear map, set by the retain-readout energy entangled with the forget-discriminant subspace. The leading source-free erasers instantiate one projection operator, and the bound predicts their retain degradation.
    \item We establish that conditioning the erasure on the input escapes this frontier (Proposition~\ref{prop:escape}). A single gate family carries explicit retain-cost and leakage bounds priced by its false-positive and missed-forget masses, and as the head separates forget from retain, its hard limit approaches the zero-cost deletion no fixed linear map can reach where the frontier is positive.
    \item We instantiate the escape as \textbf{SCOPE}, a closed-form source-free class unlearner that reads its gate from the frozen head and uses no retain data or gradient training.
    \item We demonstrate across five object, face, and speaker benchmarks on ResNet-18 and Swin-T that the frontier predicts the measured retain cost and that SCOPE leads the source-free erasers under linear and converged re-extraction, with nonlinear and sparse-dictionary audits confirming the deletion.
\end{itemize}

\section{Related Work}
\label{sec:related}

From-scratch retraining on the retain set is the reference~\citep{bourtoule2021machine,neel2021descent,guo2020certified,nguyen2025survey}, but its cost~\citep{izzo2021approximate,vanwaerebeke2025whentoforget} leads most methods to edit the trained model through gradient ascent, relabeling, retain-set finetuning, saliency masking, or distillation~\citep{rl,ftorng,salun,badt,kurmanji2023scrub,foster2024ssd,chen2023boundary,foster2024lossfree,jung2025opc}. A class quieted at the output can still survive in the features~\citep{xiao2025reminiscence,suppression,kim2025truly}, so deletion is judged at the representation level by re-extraction, adopted from~\citep{esc} and strengthened with converged, nonlinear, and sparse-dictionary probes. Our frontier concerns erasers with a fixed feature map. The closest, recovering a forget subspace or projecting off a class-weight direction, instantiate the single input-independent projection of Definition~\ref{def:global}~\citep{esc,pour,inlp}, while erasers that edit weights across layers, cancel a concept in a kernel, or act in a closed-form parameter space~\citep{kodge2024deep,ravfogel2022kernelized,zhang2026doubleprojection} fall under the same limit where their deployed feature map reduces to a fixed linear one, learned or oblique, but not otherwise (Theorem~\ref{thm:frontier}). These inherit a bound earlier work left implicit, whereas distillation lies outside it~\citep{delete}; many also need the retain set or gradient training, neither available in our source-free setting. Conditioning the erasure on the input is the route beyond this class. It already appears in language and multimodal models that withhold a detected forget query's output or steer activations only on forget concepts~\citep{guard,mllmeraser,cast}, and in a concurrent vision approach routing features onto retained-class prototypes~\citep{more}. None derives its conditioning from a deletion limit; we motivate it by the frontier and prove one gate family escapes it (Proposition~\ref{prop:escape}).

\section{Theoretical Analysis}
\label{sec:theory}

Let $\vz=\phi(\vx)\in\R^{d}$ be the penultimate feature from the frozen feature map $\phi$, the input to the frozen linear head $\W\in\R^{C\times d}$. Its rows $\vw_{c}^{\top}$ give the class weight scores $\vw_{c}^{\top}\vz$ over $C$ classes with label $Y$. The deployed head adds a learned bias; it cancels in the retain-logit perturbations and does not enter the gate, so we omit it. The classes split into a nonempty forget set $\calF$ and a retain set $\calR$ with $\Pr[Y\in\calR]>0$. Write $K=|\calF|\ge1$ for the number of forget classes, and let $\W_{\calF}$ and $\W_{\calR}$ be the forget and retain rows of $\W$. The retain-readout subspace is $\SR=\range(\W_{\calR}^{\top})$. Under the shared-covariance forget model below, let $\vmu_c=\E[\vz\mid Y{=}c]$ be the class-conditional feature means and $\bar{\vmu}_{\calF}=\frac1K\sum_{c\in\calF}\vmu_c$ their forget-set average; the forget-discriminant subspace is $\UF=\spanop\{\vmu_c-\bar{\vmu}_{\calF}:c\in\calF\}$. Its dimension $\dim\UF\in\{0,\ldots,K-1\}$ equals $K-1$ exactly when the population forget-class means are affinely independent, and $0$ only when $K=1$ or they coincide. Finally, $\Proj_{\mathcal A}$ denotes the orthogonal projector onto a subspace $\mathcal A$, so $\PUF$ projects onto $\UF$; see the appendix for full notation.

\begin{definition}[Global linear feature-projection unlearner]
\label{def:global}
A global projection unlearner maps $\vz\mapsto\Proj\vz$ at inference, where $\Proj=\I-\PcQ$, $\PcQ=\Q\Q^{\top}$, $\Q\in\R^{d\times q}$ with $0\le q\le d$, $\Q^{\top}\Q=\I_q$, and $\cQ=\range(\Q)$, one fixed projection applied to every input. We allow $q=0$, with $\Q$ the $d\times0$ matrix, $\PcQ=\Zero$, and $\cQ=\{\vzero\}$, so zero-dimensional erased subspaces are covered by the same definition.
\end{definition}

\paragraph{Deletion criterion.}
The standard deletion audit in source-free class unlearning is frozen-feature linear-probe re-extraction (knowledge-retention, KR); we adopt its worst-case population form under a shared-covariance model.
\begin{assumption}[Class-conditional model]
\label{ass:gauss}
For each class $c$, $\vz\mid Y{=}c\sim\mathcal N(\vmu_{c},\Sig)$ with shared $\Sig\succ0$ and $\Pr(Y{=}c)>0$ for every $c\in\calF\cup\calR$.
\end{assumption}
\begin{definition}[Deletion]
\label{def:delete}
Fix a bijection $\iota:\calF\to\{1,\dots,K\}$. A measurable transform $T:\R^d\to\R^d$ deletes the forget classes if every linear probe $(\Hmat,\vb)$ with $\Hmat\in\R^{K\times d}$ and $\vb\in\R^{K}$ attains forget-class balanced accuracy (denoted by $\mathrm{BA}_{\calF}(\cdot)$) at most $\frac1K$ (chance) on $\vz\mapsto\argmax_{c\in\calF}(\Hmat\,T(\vz)+\vb)_{\iota(c)}$ over forget inputs, with a fixed deterministic tie-break.
\end{definition}
This is a within-forget criterion. Full-class re-extraction remains the primary deletion standard at every $K$. When $K\ge2$, the forget classes also have between-class structure, which the within-forget criterion isolates and the frontier prices exactly. Under Assumption~\ref{ass:gauss}, a linear transform $T(\vz)=\Tmat\vz$ deletes precisely when $\MI(Y_{\calF};T(\vz))=0$, where $\MI(\cdot;\cdot)$ denotes mutual information and $Y_{\calF}$ is $Y$ conditioned on $Y\in\calF$. This classifier-independent population criterion is equivalent to $T$ annihilating $\UF$. For a global projection $\Proj=\I-\PcQ$, it reads $\UF\subseteq\cQ$ (in the appendix). Deletion fixes which subspace must go; the next question is how much retain-readout energy removing it costs.

\paragraph{Retain cost.}
Let $\mathcal D_{\calR}$ and $\mathcal D_{\calF}$ be the class-balanced feature distributions of retain-class and forget-class inputs, each an equal-weight mixture over the classes in $\calR$ resp.\ $\calF$, matching the balanced-accuracy convention. We measure damage to the retained task by the input-averaged squared perturbation of retain logits,
\begin{equation}
\label{eq:rc-main}
  \RC(T)=\sum_{r\in\calR}\E_{\vz\sim\mathcal D_{\calR}}
  \big[(\vw_{r}^{\top}\vz-\vw_{r}^{\top}T(\vz))^{2}\big].
\end{equation}
\begin{assumption}[Isotropic retain second moment]
\label{ass:white}
The retain second moment is isotropic, $\Sig_{\calR}:=\E_{\vz\sim\mathcal D_{\calR}}[\vz\vz^{\top}]=\I$.
\end{assumption}
Under Assumption~\ref{ass:white} (general retain second moment in the appendix), for a global projection \eqref{eq:rc-main} collapses to the retain-readout energy captured by the erased subspace, $\RC(\Proj)=\norm{\PcQ\W_{\calR}^{\top}}_{F}^{2}$ (derivation in the appendix); Section~\ref{sec:exp-ablation} measures this deployed cost against explicit whitening. It takes values in $[0,+\infty]$ for a general measurable $T$. Throughout, $\W$ is the deployed head's weight matrix.

\paragraph{The entanglement frontier.}
A fixed eraser removes the same directions from every input, so it cannot delete $\UF$ without disturbing the retain readout aligned with it. The deletion constraint $\UF\subseteq\cQ$ and the cost $\norm{\PcQ\W_{\calR}^{\top}}_{F}^{2}$ pull against each other, and the best trade-off is fixed by the geometry of $\UF$ and $\SR$.

\begin{theorem}[Entanglement frontier]
\label{thm:frontier}
Under Assumptions~\ref{ass:gauss} and~\ref{ass:white}, let $p=\min(\dim\UF,\dim\SR)$. Let $\theta_{1}\le\cdots\le\theta_{p}$ be the principal angles between $\UF$ and $\SR$, with corresponding $\SR$-side principal directions $\vu_{1},\dots,\vu_{p}$, unit vectors in $\SR$. Let $\rho_i=\norm{\W_{\calR}\vu_i}_2^2\ge0$ be the retain-readout energy along $\vu_i$. Under a repeated principal angle the individual $\rho_i$ depend on the chosen principal basis, but the $\cos^2\theta_i$-weighted block sum, hence the total, is invariant. Among all global projections that delete the forget classes, with the empty sum read as $0$ when $p=0$,
\begin{equation}
\label{eq:frontier-main}
  \min_{\Proj:\,\UF\subseteq\cQ}\RC(\Proj)
  =\norm{\PUF\W_{\calR}^{\top}}_{F}^{2}
  =\sum_{i=1}^{p}\cos^{2}\theta_{i}\,\rho_{i}.
\end{equation}
The minimum is attained by the minimal eraser $\cQ=\UF$. The minimizers are exactly the feasible erasers with $\cQ\ominus\UF\perp\SR$, writing $\cQ\ominus\UF:=\cQ\cap\UF^{\perp}$, hence unique up to $\SR$-orthogonal padding. Every deleting linear map pays at least this floor, since $\UF\subseteq\ker(\Tmat)$ forces $\RC(\Tmat)\ge\norm{\PUF\W_{\calR}^{\top}}_{F}^{2}$, so~\eqref{eq:frontier-main} is the exact minimum over all deleting linear maps. The minimum is zero iff $\UF\perp\SR$.
\end{theorem}

\begin{corollary}[General $\Sig_{\calR}$]
\label{cor:general}
Under Assumption~\ref{ass:gauss} and for any $\Sig_{\calR}\succ\Zero$, the coordinates $\vz\mapsto\Sig_{\calR}^{-\frac{1}{2}}\vz$ make the retain second moment isotropic. Theorem~\ref{thm:frontier} then gives $\norm{\Proj_{\Sig_{\calR}^{-\frac{1}{2}}\UF}(\W_{\calR}\Sig_{\calR}^{\frac{1}{2}})^{\top}}_{F}^{2}$ as the exact minimum over all deleting linear maps (in the appendix).
\end{corollary}

\paragraph{Escaping the frontier.}
When the frontier value~\eqref{eq:frontier-main} is positive, no fixed linear eraser, projection or not, deletes at zero retain cost. A fixed map applies the same erasure to every input. Conditioning instead lets the erasure strength depend on the input, so its retain cost can scale with the gate's mass on retain inputs. For a measurable gate $g:\R^{d}\to[0,1]$ and any erased subspace $\cQ$, define the input-conditional map
\begin{equation}
\label{eq:cond-main}
  M_{g}(\vz)=\vz-g(\vz)\,\PcQ\vz .
\end{equation}
When $\PcQ\ne0$ and $g$ is nonconstant on $\{\PcQ\vz\ne\vzero\}$, $M_{g}$ is nonlinear, hence neither a fixed projection nor a linear map.

\begin{proposition}[Conditional escape]
\label{prop:escape}
Assume Assumption~\ref{ass:gauss} and let $\cQ$ be any erased subspace. Consider the conditional map~\eqref{eq:cond-main}. Let $M_h$ use a measurable hard gate $h:\R^d\to\{0,1\}$, and let $M_g$ use a measurable soft gate $g:\R^d\to[0,1]$. On the retain side, $\alpha_h=\Pr_{\mathcal D_{\calR}}[h(\vz){=}1]$ is the false-positive rate and $\alpha=\E_{\mathcal D_{\calR}}[g(\vz)]$ the soft retain mass. On the forget side, $\bar\epsilon=\frac1K\sum_{c\in\calF}\Pr[h(\vz){=}0\mid Y{=}c]$ is the average miss rate and $\beta=\E_{\mathcal D_{\calF}}[1-g(\vz)]$ the missed-forget mass. Write $\Leak(M_g)=\E_{\mathcal D_{\calF}}\norm{\PUF M_g(\vz)}_2^2$ for the residual forget-discriminant energy, and $\kappa_{\calR}=\sum_{r\in\calR}(\E_{\mathcal D_{\calR}}[(\vw_r^\top\PcQ\vz)^4])^{\frac12}$, $\kappa_{\calF}=(\E_{\mathcal D_{\calF}}\norm{\PUF\vz}_2^4)^{\frac12}$ for the finite gate-independent fourth-moment constants. The retain-cost bounds hold for any $\cQ$; if $\UF\subseteq\cQ$ the deletion and leakage bounds also hold, the former for every measurable classifier $\psi:\R^d\to\calF$ and hence every linear probe:
\[
\begin{aligned}
  \RC(M_h)&\le\sqrt{\alpha_h}\,\kappa_{\calR}, & \RC(M_g)&\le\sqrt{\alpha}\,\kappa_{\calR},\\
  \mathrm{BA}_{\calF}(\psi\circ M_h)&\le\tfrac1K+\bar\epsilon, & \Leak(M_g)&\le\sqrt{\beta}\,\kappa_{\calF}.
\end{aligned}
\]
If $\UF\subseteq\cQ$, a gate firing exactly on forget inputs ($\bar\epsilon{=}\alpha_h{=}0$) attains exact deletion at $\RC(M_h){=}0$. When the frontier is positive, no fixed deleting linear map reaches this zero-cost corner. The hard map approaches it as the head's weight scores separate forget from retain. The appendix gives the limiting statement and fixed-instance behavior.
\end{proposition}

\section{Spectral Conditional Projective Erasure}
\label{sec:scope}

Spectral Conditional Projective Erasure (SCOPE) instantiates the conditional map~\eqref{eq:cond-main} with two components: an orthonormal erased basis $\Q$ ($\cQ=\range(\Q)$) and a head-derived gate. The basis fixes which directions are erased, and the gate where and how strongly. Both are built in closed form from the forget features and the frozen head, with no retain data or gradient training. Algorithm~\ref{alg:scope} summarizes the construction. Proposition~\ref{prop:escape}'s retain-cost bound needs no containment and applies to deployed SCOPE. Its deletion certificate and leakage bound hold under the population containment $\UF\subseteq\cQ$, and KR re-extraction audits deletion at finite $\tau$ and finite samples.

\paragraph{Erased basis.}
SCOPE's erased basis augments the empirical forget-discriminant subspace with residual forget directions. For each forget class $c\in\calF$, let $\mathcal X_c$ be its nonempty sample set and $\hat\vmu_c=\frac{1}{|\mathcal X_c|}\sum_{\vx\in\mathcal X_c}\phi(\vx)$ its empirical feature mean. Write $\mathcal X_{\calF}=\bigcup_{c\in\calF}\mathcal X_c$ and $\mathbf{Z}_{\calF}=[\phi(\vx)]_{\vx\in\mathcal X_{\calF}}\in\R^{d\times|\mathcal X_{\calF}|}$. The empirical forget-discriminant subspace is $\hUF=\spanop\{\hat\vmu_c-\hat\vmu_{c'}:c,c'\in\calF\}=\range(\hBF)$, where $\hBF^{\top}\hBF=\I$ and $\hat s_{\calF}:=\dim\hUF\in\{0,\ldots,K-1\}$. When $\hat s_{\calF}=0$, $\hBF$ is the empty $d\times0$ matrix and $\hBF\hBF^{\top}=\Zero$. In the setting of Theorem~\ref{thm:frontier}, the population deletion target is $\UF$ and each erased unit direction $\vv$ is priced by the per-direction retain-cost functional $\operatorname{cost}_{\calR}(\vv)=\norm{\W_{\calR}\vv}_2^2$. This head-space quantity is available from the frozen head without retain data. The appendix treats the general retain second moment. Summing $\operatorname{cost}_{\calR}$ over an orthonormal basis of $\range(\Q)$ recovers the global cost $\norm{\PcQ\W_{\calR}^{\top}}_F^2$ that Theorem~\ref{thm:frontier} minimizes.

\begin{algorithm}[t]
\caption{Spectral Conditional Projective Erasure}
\label{alg:scope}
\begin{algorithmic}[1]
\renewcommand{\algorithmicrequire}{\textbf{Input:}}
\renewcommand{\algorithmicensure}{\textbf{Output:}}
\REQUIRE frozen feature map $\phi$ and head $\W$ (forget/retain rows $\W_{\calF},\W_{\calR}$); forget/retain classes $\calF,\calR$; nonempty forget samples $\{\mathcal X_c\}_{c\in\calF}$
\renewcommand{\algorithmicrequire}{\textbf{Parameter:}}
\REQUIRE $K$-adaptive erased-rank schedule $q_{\mathrm{sched}}(\cdot)\in\mathbb{Z}_{\ge0}$; residual candidate pool $r_{\mathrm{pool}}\in\mathbb{Z}_{\ge0}$ with $r_{\mathrm{pool}}\ge q_{\mathrm{sched}}(|\calF|)$; gate scale $\tau\in(0,\infty]$
\ENSURE erased feature map $M:\R^{d}\to\R^{d}$

\STATE $\mathbf{Z}_{\calF}\leftarrow[\phi(\vx)]_{\vx\in\mathcal X_{\calF}}$,\quad $\hat\vmu_c\leftarrow\frac{1}{|\mathcal X_c|}\sum_{\vx\in\mathcal X_c}\phi(\vx)$
\STATE $\hBF\leftarrow$ basis of $\hUF=\spanop\{\hat\vmu_c-\hat\vmu_{c'}:c,c'\in\calF\}$;\quad $\hat s_{\calF}\leftarrow\dim\hUF$ \COMMENT{forget-discriminant subspace}
\STATE $\mathbf{R}\leftarrow(\I-\hBF\hBF^{\top})\mathbf{Z}_{\calF}$ \COMMENT{residual off $\hUF$}
\STATE $\vv_{(1)},\vv_{(2)},\dots\leftarrow$ the $\min\{r_{\mathrm{pool}},\rank\mathbf{R}\}$ eigenvectors of $\mathbf{R}\mathbf{R}^{\top}$ with the largest eigenvalues $\lambda_j\!>\!0$, re-sorted by decreasing $\omega$ \COMMENT{Eq.~\eqref{eq:scope-score}}
\STATE $q\leftarrow\min\{\max(q_{\mathrm{sched}}(|\calF|),\hat s_{\calF}),\rank\mathbf{Z}_{\calF}\}$
\STATE $\Q\leftarrow[\,\hBF\mid\vv_{(1)},\dots,\vv_{(q-\hat s_{\calF})}\,]$ \COMMENT{Eq.~\eqref{eq:scope-basis}}
\STATE \textbf{return} $M(\vz)=\vz-g_{\tau}(\vz)\,\Q\Q^{\top}\vz$ \COMMENT{Eqs.~\eqref{eq:cond-main} and~\eqref{eq:scope-gate}}
\end{algorithmic}
\end{algorithm}

Writing $\mathbf{R}=(\I-\hBF\hBF^{\top})\mathbf{Z}_{\calF}$, define the retain-cost-normalized forget-readout score
\begin{equation}\label{eq:scope-score}
  \omega(\vv)=\frac{\norm{\W_{\calF}\vv}_2^2}{\operatorname{cost}_{\calR}(\vv)}\in[0,+\infty],
\end{equation}
read as $+\infty$ when $\operatorname{cost}_{\calR}(\vv)=0<\norm{\W_{\calF}\vv}_2$ and as $0$ when $\operatorname{cost}_{\calR}(\vv)=\norm{\W_{\calF}\vv}_2=0$. A larger $\omega(\vv)$ signals more forget readout per unit retain cost. Let $\vv_{(1)},\vv_{(2)},\dots$ be the eigenvectors of $\mathbf{R}\mathbf{R}^{\top}$ with nonzero eigenvalue, in a fixed orthonormal basis within any repeated eigenspace. They lie in $\hUF^{\perp}$, $[\,\hBF\mid\vv_{(1)},\dots\,]$ is orthonormal, and their number is $\rank\mathbf{Z}_{\calF}-\hat s_{\calF}$. The candidate set consists of the $r_{\mathrm{pool}}$ eigenvectors with the largest eigenvalues, or all of them when fewer remain (in the appendix). We order this candidate set by descending score $\omega(\vv_{(1)})\ge\omega(\vv_{(2)})\ge\cdots$, breaking ties by a deterministic rule. With the basis within each repeated eigenspace fixed, the construction is reproducible. If the selected rank truncates an exactly degenerate eigenspace, $\range(\Q)$ depends on that fixed basis. Let $q_{\mathrm{sched}}(K)$ be the $K$-adaptive erased-rank schedule in the appendix. It increases with $K$ to accommodate the larger possible forget-discriminant dimension and residual re-extractable energy. SCOPE sets the erased rank $q=\min\{\max(q_{\mathrm{sched}}(K),\hat s_{\calF}),\rank\mathbf{Z}_{\calF}\}$, so that $\hat s_{\calF}\le q\le\rank\mathbf{Z}_{\calF}$. Taking $r_{\mathrm{add}}=q-\hat s_{\calF}\le q_{\mathrm{sched}}(K)\le r_{\mathrm{pool}}$ residual directions, SCOPE erases
\begin{equation}
\label{eq:scope-basis}
  \range(\Q)=\hUF\oplus\spanop\{\vv_{(1)},\dots,\vv_{(r_{\mathrm{add}})}\}.
\end{equation}
When $r_{\mathrm{add}}=0$ the residual span is $\{\vzero\}$ and $\Q=\hBF$. By~\eqref{eq:scope-basis} $\hUF\subseteq\range(\Q)$. In the population construction, replacing empirical by population means gives $\hUF=\UF$, hence $\UF\subseteq\range(\Q)$, the containment required by the deletion and leakage bounds of Proposition~\ref{prop:escape}. The minimal eraser $\UF$ is sufficient (shown in the appendix) and retain-cost optimal among deleting projections (Theorem~\ref{thm:frontier}) in the population linear model. The residual block targets finite-sample forget energy still re-extractable beyond $\hUF$.

\paragraph{Gate family.}
The forget-minus-retain margin is $m(\vz)=\max_{c\in\calF}(\W\vz)_c-\max_{c\in\calR}(\W\vz)_c$. With the logistic $\sigma(t)=\frac{1}{1+e^{-t}}$, SCOPE's gate is a family, read off the frozen head's weight rows and indexed by a scale $\tau\in(0,\infty]$:
\begin{equation}
\label{eq:scope-gate}
  g_{\tau}(\vz)=\sigma\big(\tau\,m(\vz)\big).
\end{equation}
Thus $m(\vz)>0$ when the largest forget-class weight score exceeds the largest retain-class one, and $\tau$ controls the gate's sharpness. At the endpoint, $g_{\infty}(\vz):=\mathbf 1\{m(\vz)>0\}$. Under Assumption~\ref{ass:gauss} and distinct cross-split rows, the tie event $m(\vz)=0$ has probability zero, and $g_\tau$ converges almost surely as $\tau\to\infty$ to the hard indicator gate $h(\vz)=\mathbf 1\{\argmax_j(\W\vz)_j\in\calF\}$, while every finite $\tau$ is a soft member of Proposition~\ref{prop:escape}. The two ends' rates are compared in the appendix. Because $m$ is a difference of weight scores, $\tau$ is relative to their scale ($\W\mapsto a\W$ with $a>0$ sends $g_\tau\mapsto g_{a\tau}$). The $\tau\to\infty$ hard limit is scale-free, and re-extraction is empirically flat across $\tau$ (see the appendix). When $\PcQ\ne0$ and $g_\tau$ is nonconstant on $\{\PcQ\vz\ne\vzero\}$, the deployed map is nonlinear and lies outside the fixed-linear class of Theorem~\ref{thm:frontier}. The deployed finite-$\tau$ gate meets this nondegeneracy under a full-support law with distinct cross-split rows and $\UF\ne\{\vzero\}$.

\section{Experiments}
\label{sec:exp}

\subsection{Experimental Setup}

\paragraph{Benchmarks, backbones, and methods.}
We evaluate source-free class unlearning across two modalities, the image classifiers CIFAR-100~\citep{cifar100}, Tiny-ImageNet~\citep{tinyimagenet} (200 classes), and the VGGFace2~\citep{vggface2} subsets VGGFace2-100 and VGGFace2-200, and the speaker benchmark VCTK~\citep{vctk}. The backbones are ResNet-18~\citep{resnet} and Swin-T~\citep{swin}, applied to log-mel spectrograms for VCTK. For every benchmark and $K\in\{1,2,5,10,20\}$ we average three forget-set trials; dataset, forget-set, hyperparameter-search, and evaluation details are in the appendix (with the deployed configuration and selection rules). Beyond Original and the from-scratch Retrain on $\calR$, we compare the forget-data-only unlearners ESC~\citep{esc} and POUR-P~\citep{pour} (the projection erasers closest to SCOPE), DELETE~\citep{delete}, NG~\citep{ftorng}, and RL~\citep{rl}, and the retain-data unlearners FT~\citep{ftorng}, SalUn~\citep{salun}, and BadT~\citep{badt}, each trained baseline tuned per setting.

\paragraph{Metrics and protocol.}
Every audit probes what each method deploys, the deployed-representation standard of source-free class unlearning; for SCOPE that is the frozen model with feature map $M\circ\phi$, audited exactly as the erasers it is compared against. Splitting every dataset into retain/forget train ($D_r,D_f$), validation, and test ($D_{rt},D_{ft}$), we report accuracy on each split with the frozen deployed head and under knowledge-retention (KR) re-extraction, a full-class linear probe retrained on the unlearned training features, the standard audit of~\citep{esc}. The validation split serves only baseline hyperparameter and rank selection; every test-split metric is computed once, and the train-side panels ($D_f$, $D_r$, and $HM$) are split-independent. We summarize deletion against retention with the harmonic mean $HM$ of the retain accuracy and the forget error $100{-}D_{f}$ (reusing the split symbols for the accuracies; $HM$ is computed per trial before averaging), and its test-split analogue $HM_{t}$, superscripted N (NORMAL, deployed head) or KR (re-extraction) where the panel matters. Our primary evidence against residual leakage is the re-extraction ladder, strengthened from the linear KR probe to a converged L-BFGS~\citep{liu1989limited} re-extractor (in the appendix) and a nonlinear MLP. We also report a support-vector-classifier membership-inference attack on the true-class confidences (MIA)~\citep{mia}. We report the forget separability $S_{\calF}$, the balanced accuracy of a $K$-way linear probe over the forget classes' training features (chance $\frac{100}{K}$), a trained-probe lower bound on the within-forget deletion criterion of Section~\ref{sec:theory}. These form three deletion standards, ordered by scope and probe capacity, the within-forget criterion ($S_{\calF}$, undefined at $K{=}1$), the broader full-class KR re-extraction, the primary standard, and the nonlinear retraining-relative standard in the appendix (recover no more than the retraining reference). A single forget class carries no between-class geometry, so the within-forget criterion and frontier are $K{\ge}2$ statements, while single-class forgetting is judged by full-class re-extraction, the standard at every $K$, on which SCOPE drives forget recovery to near zero. A sparse-autoencoder (SAE) audit on the image benchmarks reaches the same verdict at the sparse-dictionary level (in the appendix).

\subsection{The Frontier Predicts Retain Cost}
\label{sec:exp-frontier}

\begin{figure}[t]
\centering
\includegraphics[width=\columnwidth]{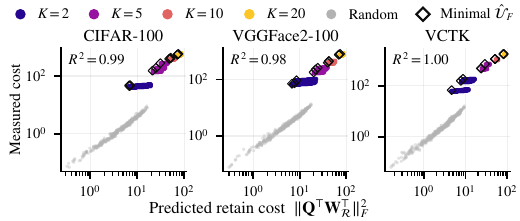}
\caption{The entanglement frontier predicts retain cost (CIFAR-100, VGGFace2-100, and VCTK on Swin-T; all ten pairs in the appendix). The geometry-only prediction of Theorem~\ref{thm:frontier} versus the measured retain perturbation of generic deleting projections at each size $K$ and of matched-rank random subspaces that do not delete; protocol in the appendix.}
\label{fig:predictivity}
\end{figure}

The geometry-only prediction of Theorem~\ref{thm:frontier} tracks the measured retain cost across every benchmark, backbone, and forget-set size, at $R^{2}$ between $0.93$ and $1.00$ (Figure~\ref{fig:predictivity}). Sampling generic deleting projections, the minimal eraser $\hUF$ is almost always the cheapest, and matched-rank random subspaces that do not delete cost far less to erase, a gap widening with $K$ and the entanglement of Figure~\ref{fig:entangle}. The cost is entanglement, not rank, and tracks the geometry even in deployed coordinates, a stronger statement than the theorem guarantees, whose exact general-$\Sig_{\calR}$ form is Corollary~\ref{cor:general}. The published ESC falls on the curve at a median $R^{2}$ of $0.99$, and the geometry tracks the end-to-end accuracy drop where entanglement is strongest at Spearman $0.92$ (in the appendix).

\subsection{The Gate Escapes the Frontier}
\label{sec:exp-escape}

\begin{figure}[t]
\centering
\includegraphics[width=\columnwidth]{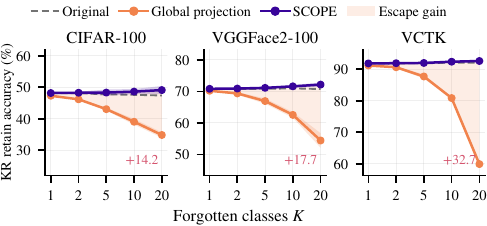}
\caption{The gate escapes the frontier (CIFAR-100, VGGFace2-100, and VCTK on Swin-T; all five in the appendix). With the erased basis $\Q$ fixed, the global projection ($g\equiv1$) shows the retention loss associated with the frontier while SCOPE's gated map stays at the original level (KR re-extraction; shading marks the escape gain).}
\label{fig:escape}
\end{figure}

With the erased basis $\Q$ frozen so the global projection $\vz-\Q\Q^{\top}\vz$ and the gated map $M$ differ only by the gate (Figure~\ref{fig:escape}), SCOPE recovers up to $32.7$ points of re-extraction retain accuracy on the most entangled pair, the escape gain tracking the frontier. This gain is due to the gate, since across all ten pairs at $K{=}20$ the deployed gate's soft retain mass never exceeds $8.9\%$ while the global map projects every input, the scaling Proposition~\ref{prop:escape} predicts (in the appendix). In every setting the measured $\RC(M_{g})$ and $\Leak(M_{g})$ stay at least twice inside $\sqrt{\alpha}\,\kappa_{\calR}$ and $\sqrt{\beta}\,\kappa_{\calF}$, with the fourth-moment constants $\kappa_{\calR},\kappa_{\calF}$ estimated from the features, and $\RC(M_{g})$ sits below the general-$\Sig_{\calR}$ frontier floor of Corollary~\ref{cor:general} in every trial (in the appendix) and across the larger forget sets $K\in\{5,10,20\}$, by $2.3$ to $\sim299\times$ (in the appendix). The deployed head needs no calibration; a binary detector shows no increased black-box forget-region detectability (in the appendix), and under distribution shift retention holds at the original level where the global projection collapses (in the appendix). The escape costs no deletion, since both arms drive linear forget recovery to near zero and the gated map stays at or below the retraining reference under a nonlinear probe in most settings.

\subsection{Benchmark Comparison}
\label{sec:exp-benchmark}

\begin{table}[t]
\centering
\setlength{\tabcolsep}{1.0pt}
\renewcommand{\arraystretch}{1.05}
\begingroup
\tablefontsize
\begin{tabular}{@{}l c c c c c@{}}
\toprule
\multirow{2}{*}{Method} & \multicolumn{5}{c}{$HM_t^{\mathrm{KR}}$ ($\uparrow$) at forget-set size $K$} \\
\cmidrule(lr){2-6}
 & $1$ & $2$ & $5$ & $10$ & $20$ \\
\midrule
Original & \tabpm{11.3}{5.9} & \tabpm{16.5}{11.9} & \tabpm{16.1}{4.5} & \tabpm{19.2}{5.8} & \tabpm{16.5}{1.9} \\
Retrain & \tabpm{14.1}{5.2} & \tabpm{27.8}{18.8} & \tabpm{31.9}{7.7} & \tabpm{35.7}{3.3} & \tabpm{29.7}{1.9} \\
\midrule
ESC & \secondpm{95.4}{0.1} & \secondpm{95.1}{0.2} & \secondpm{93.4}{0.4} & \secondpm{90.5}{0.1} & \secondpm{82.1}{1.7} \\
POUR-P & \tabpm{84.2}{5.5} & \tabpm{79.3}{11.1} & \tabpm{84.1}{4.6} & \tabpm{85.5}{5.6} & \tabpm{80.1}{1.6} \\
DELETE & \tabpm{21.2}{6.6} & \tabpm{26.3}{15.1} & \tabpm{25.5}{3.3} & \tabpm{29.9}{7.1} & \tabpm{27.5}{2.5} \\
\textbf{SCOPE} & \bestpm{95.8}{0.0} & \bestpm{95.8}{0.1} & \bestpm{95.8}{0.1} & \bestpm{96.0}{0.2} & \bestpm{95.9}{0.2} \\
\bottomrule
\end{tabular}
\endgroup
\caption{$HM_t^{\mathrm{KR}}$ across forget-set sizes on VCTK (Swin-T); it is the metric comparable across $K$, cells are mean$\pm$std over trials, best unlearner \textbf{bold}, second \underline{underlined}.}
\label{tab:byk}
\end{table}

On the most entangled pair SCOPE's re-extraction score stays flat across forget-set sizes while the global ESC decays with $K$, so their gap widens (Table~\ref{tab:byk}). Table~\ref{tab:main} takes the hardest point, $K{=}20$, across four benchmark and backbone pairs spanning two modalities and three recognition domains. There the deployed panel saturates for the leading methods, collapsing their differences, so re-extraction separates them, and SCOPE tops the unlearners by deleting what the baselines leave re-extractable, its re-extraction $HM_t^{\mathrm{KR}}$ best on all four pairs and above even retraining ($97.4$ against $33.5$ on VGGFace2-200). POUR-P and the trained methods keep the forget classes recoverable, and ESC removes them only partway, worst where entanglement is highest. Even retraining leaves the forgotten classes linearly separable (also on held-out features, in the appendix), whereas placing the empirical forget-discriminant subspace inside the erased basis and suppressing it collapses that separability toward chance. Deletion here therefore targets unrecoverability from deployed features, and re-extraction below the retraining reference is the intended privacy guarantee. A converged linear adversary rules out a probe-budget artifact, SCOPE's forget recovery lowest in all twenty rows at $K{\in}\{10,20\}$ (in the appendix). At $K{=}20$ an MLP ladder reaching $1024$ units holds recovery at or below the retraining reference at $46$ of the $50$ rungs, the four exceptions near-saturated on VGGFace2 (ResNet-18), and well below it on Swin-T, the reference for what a probe re-learns without stored forget information (in the appendix). Figure~\ref{fig:teaser} shows this geometrically and Figure~\ref{fig:retrieval} at the task level, with more retrieval examples and per-setting grids in the appendix. The edit is $10^{4}$ to $10^{5}\times$ faster than retraining (in the appendix), its forget eigendecomposition $3.2\times$ to $6.4\times$ faster with a randomized solver (in the appendix), and its erased rank a fixed function of $K$ alone (in the appendix).

\begin{table*}[t]
\centering
\setlength{\tabcolsep}{1.8pt}
\renewcommand{\arraystretch}{1.05}
\begingroup
\tablefontsize
\begin{tabular}{@{}l c c c c c c c c c c c @{}}
\toprule
\multirow{2}{*}{Method} & \multirow{2}{*}{Retain} & \multicolumn{5}{c}{CIFAR-100 (ResNet-18)} & \multicolumn{5}{c}{Tiny-ImageNet (Swin-T)} \\
\cmidrule(lr){3-7}\cmidrule(lr){8-12}
 &  & $HM^{\mathrm{N}}\,\uparrow$ & $HM_t^{\mathrm{N}}\,\uparrow$ & MIA$\,\downarrow$ & $HM_t^{\mathrm{KR}}\,\uparrow$ & $S_{\calF}\,\downarrow$ & $HM^{\mathrm{N}}\,\uparrow$ & $HM_t^{\mathrm{N}}\,\uparrow$ & MIA$\,\downarrow$ & $HM_t^{\mathrm{KR}}\,\uparrow$ & $S_{\calF}\,\downarrow$ \\
\midrule
Original & -- & \tabpm{0.1}{0.0} & \tabpm{31.5}{2.6} & \tabpm{99.1}{0.0} & \tabpm{34.8}{2.4} & \tabpm{100.0}{0.0} & \tabpm{35.5}{1.8} & \tabpm{47.4}{1.7} & \tabpm{79.0}{1.3} & \tabpm{46.1}{0.8} & \tabpm{88.8}{2.0} \\
Retrain & -- & \tabpm{100.0}{0.0} & \tabpm{88.5}{0.4} & \tabpm{0.0}{0.0} & \tabpm{83.8}{0.6} & \tabpm{76.5}{1.6} & \tabpm{85.1}{0.5} & \tabpm{67.3}{0.3} & \tabpm{0.0}{0.0} & \tabpm{51.4}{0.5} & \tabpm{68.1}{2.1} \\
\midrule
FT & \multirow{3}{*}{$\checkmark$} & \tabpm{99.8}{0.1} & \tabpm{86.8}{0.7} & \bestpm{0.0}{0.0} & \tabpm{63.9}{2.9} & \tabpm{94.6}{0.9} & \bestpm{86.8}{0.9} & \tabpm{65.5}{0.4} & \tabpm{3.4}{1.3} & \tabpm{49.1}{0.7} & \tabpm{80.5}{2.3} \\
SalUn &  & \tabpm{99.2}{0.3} & \tabpm{86.3}{0.4} & \bestpm{0.0}{0.0} & \tabpm{79.2}{2.4} & \tabpm{91.8}{0.9} & \secondpm{86.5}{0.4} & \tabpm{65.5}{0.3} & \tabpm{1.6}{0.6} & \tabpm{49.1}{0.8} & \tabpm{80.7}{2.5} \\
BadT &  & \tabpm{98.9}{0.0} & \tabpm{83.3}{0.5} & \bestpm{0.0}{0.0} & \tabpm{83.4}{0.8} & \tabpm{96.0}{0.3} & \tabpm{74.2}{0.7} & \tabpm{64.2}{0.6} & \tabpm{2.9}{0.4} & \tabpm{48.8}{0.3} & \tabpm{86.5}{2.1} \\
\midrule
NG & \multirow{6}{*}{$\times$} & \tabpm{60.3}{17.8} & \tabpm{69.4}{3.6} & \tabpm{24.3}{10.2} & \tabpm{64.2}{8.1} & \tabpm{99.1}{1.1} & \tabpm{46.6}{4.4} & \tabpm{51.3}{2.9} & \tabpm{71.1}{5.2} & \tabpm{46.7}{0.9} & \tabpm{88.9}{2.1} \\
RL &  & \tabpm{84.3}{3.7} & \tabpm{67.7}{3.4} & \secondpm{0.4}{0.1} & \tabpm{40.0}{15.4} & \tabpm{91.0}{1.8} & \tabpm{44.3}{2.5} & \tabpm{50.5}{2.2} & \tabpm{73.7}{2.8} & \tabpm{46.6}{1.2} & \tabpm{88.9}{2.0} \\
ESC &  & \secondpm{99.9}{0.0} & \tabpm{87.5}{0.8} & \bestpm{0.0}{0.0} & \tabpm{85.0}{0.7} & \secondpm{8.1}{0.5} & \tabpm{65.8}{3.3} & \tabpm{55.8}{0.6} & \tabpm{33.4}{11.7} & \tabpm{48.8}{0.7} & \secondpm{77.6}{3.2} \\
POUR-P &  & \bestpm{100.0}{0.0} & \secondpm{88.3}{0.5} & \bestpm{0.0}{0.0} & \secondpm{85.5}{0.1} & \tabpm{99.9}{0.0} & \tabpm{80.5}{0.5} & \tabpm{62.7}{0.9} & \bestpm{0.0}{0.0} & \secondpm{52.3}{0.8} & \tabpm{88.1}{1.9} \\
DELETE &  & \tabpm{98.9}{0.3} & \tabpm{88.1}{0.5} & \bestpm{0.0}{0.0} & \tabpm{74.9}{3.6} & \tabpm{99.0}{0.2} & \tabpm{82.7}{0.5} & \secondpm{66.8}{0.4} & \tabpm{3.7}{1.8} & \tabpm{47.9}{0.6} & \tabpm{85.7}{2.5} \\
\textbf{SCOPE} &  & \bestpm{100.0}{0.0} & \bestpm{88.5}{0.4} & \bestpm{0.0}{0.0} & \bestpm{86.9}{0.5} & \bestpm{3.8}{0.3} & \tabpm{85.4}{0.1} & \bestpm{67.0}{0.2} & \secondpm{0.9}{0.2} & \bestpm{56.9}{0.2} & \bestpm{20.8}{0.3} \\
\midrule
\multirow{2}{*}{Method} & \multirow{2}{*}{Retain} & \multicolumn{5}{c}{VGGFace2-200 (Swin-T)} & \multicolumn{5}{c}{VCTK (Swin-T)} \\
\cmidrule(lr){3-7}\cmidrule(lr){8-12}
 &  & $HM^{\mathrm{N}}\,\uparrow$ & $HM_t^{\mathrm{N}}\,\uparrow$ & MIA$\,\downarrow$ & $HM_t^{\mathrm{KR}}\,\uparrow$ & $S_{\calF}\,\downarrow$ & $HM^{\mathrm{N}}\,\uparrow$ & $HM_t^{\mathrm{N}}\,\uparrow$ & MIA$\,\downarrow$ & $HM_t^{\mathrm{KR}}\,\uparrow$ & $S_{\calF}\,\downarrow$ \\
\midrule
Original & -- & \tabpm{0.0}{0.0} & \tabpm{3.9}{2.1} & \tabpm{100.0}{0.0} & \tabpm{9.4}{2.5} & \tabpm{100.0}{0.0} & \tabpm{0.8}{0.2} & \tabpm{8.6}{1.3} & \tabpm{96.2}{0.2} & \tabpm{16.5}{1.9} & \tabpm{99.8}{0.1} \\
Retrain & -- & \tabpm{100.0}{0.0} & \tabpm{98.8}{0.1} & \tabpm{0.0}{0.0} & \tabpm{33.5}{5.1} & \tabpm{89.6}{1.1} & \tabpm{99.7}{0.2} & \tabpm{97.6}{0.4} & \tabpm{0.0}{0.0} & \tabpm{29.7}{1.9} & \tabpm{98.0}{0.2} \\
\midrule
FT & \multirow{3}{*}{$\checkmark$} & \tabpm{99.7}{0.0} & \tabpm{98.2}{0.0} & \bestpm{0.0}{0.0} & \tabpm{27.8}{5.3} & \tabpm{94.2}{0.6} & \secondpm{98.7}{0.7} & \tabpm{97.0}{0.3} & \bestpm{0.0}{0.0} & \tabpm{24.9}{5.0} & \tabpm{98.5}{0.7} \\
SalUn &  & \secondpm{99.8}{0.0} & \secondpm{98.3}{0.0} & \bestpm{0.0}{0.0} & \tabpm{25.0}{3.9} & \tabpm{95.6}{0.5} & \tabpm{98.2}{0.7} & \tabpm{96.6}{0.3} & \bestpm{0.0}{0.0} & \tabpm{31.2}{3.6} & \tabpm{98.2}{0.5} \\
BadT &  & \tabpm{98.5}{0.1} & \tabpm{98.2}{0.1} & \tabpm{33.3}{57.7} & \tabpm{73.9}{4.1} & \tabpm{97.3}{0.7} & \tabpm{98.4}{0.1} & \secondpm{97.1}{0.1} & \bestpm{0.0}{0.0} & \tabpm{66.8}{4.5} & \tabpm{99.1}{0.3} \\
\midrule
NG & \multirow{6}{*}{$\times$} & \tabpm{38.2}{32.1} & \tabpm{37.4}{33.0} & \tabpm{83.0}{17.3} & \tabpm{14.3}{11.4} & \tabpm{96.8}{5.0} & \tabpm{45.0}{3.3} & \tabpm{45.6}{4.3} & \tabpm{41.0}{29.0} & \tabpm{33.1}{9.9} & \tabpm{99.0}{0.8} \\
RL &  & \tabpm{72.8}{0.0} & \tabpm{76.0}{2.2} & \tabpm{97.5}{1.3} & \tabpm{35.4}{7.7} & \tabpm{99.4}{0.3} & \tabpm{19.6}{24.7} & \tabpm{24.5}{20.8} & \tabpm{86.6}{10.6} & \tabpm{27.3}{14.8} & \tabpm{99.2}{0.8} \\
ESC &  & \tabpm{93.3}{1.2} & \tabpm{92.1}{1.7} & \secondpm{3.9}{3.7} & \secondpm{90.7}{0.4} & \secondpm{61.7}{10.5} & \tabpm{82.0}{1.6} & \tabpm{80.3}{1.7} & \tabpm{1.4}{0.7} & \secondpm{82.1}{1.7} & \secondpm{91.6}{1.0} \\
POUR-P &  & \secondpm{99.8}{0.0} & \secondpm{98.3}{0.2} & \bestpm{0.0}{0.0} & \tabpm{60.2}{2.8} & \tabpm{99.9}{0.0} & \tabpm{92.9}{0.2} & \tabpm{90.9}{0.7} & \bestpm{0.0}{0.0} & \tabpm{80.1}{1.6} & \tabpm{99.0}{0.3} \\
DELETE &  & \tabpm{99.1}{0.3} & \tabpm{97.1}{0.2} & \bestpm{0.0}{0.0} & \tabpm{22.7}{6.2} & \tabpm{97.8}{0.3} & \tabpm{97.6}{0.3} & \tabpm{95.6}{0.3} & \secondpm{0.2}{0.1} & \tabpm{27.5}{2.5} & \tabpm{99.3}{0.2} \\
\textbf{SCOPE} &  & \bestpm{100.0}{0.0} & \bestpm{98.9}{0.1} & \bestpm{0.0}{0.0} & \bestpm{97.4}{0.1} & \bestpm{0.7}{0.0} & \bestpm{99.8}{0.0} & \bestpm{97.9}{0.1} & \bestpm{0.0}{0.0} & \bestpm{95.9}{0.2} & \bestpm{1.5}{0.1} \\
\bottomrule
\end{tabular}
\endgroup
\caption{Class unlearning at $K{=}20$. Mean$\pm$std over trials; the Retain column marks retain-data use (reference rows marked --); best \textbf{bold}, second \underline{underlined} among unlearners. MIA attacks the forget training samples.}
\label{tab:main}
\end{table*}

\begin{figure}[t]
    \centering
    \includegraphics[width=\columnwidth]{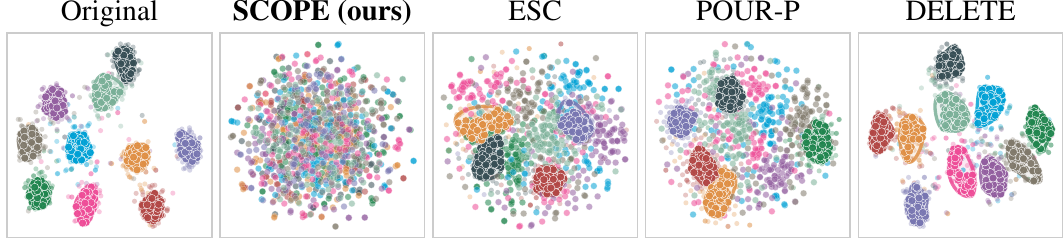}
    \caption{t-SNE~\citep{tsne} of forget-class features (VGGFace2-100, Swin-T, $K{=}10$). SCOPE's forget identities appear as one mass, while baselines keep them clustered, leaving identity structure at the feature level.}
    \label{fig:teaser}
\end{figure}

\begin{figure}[t]
    \centering
    \includegraphics[width=\columnwidth]{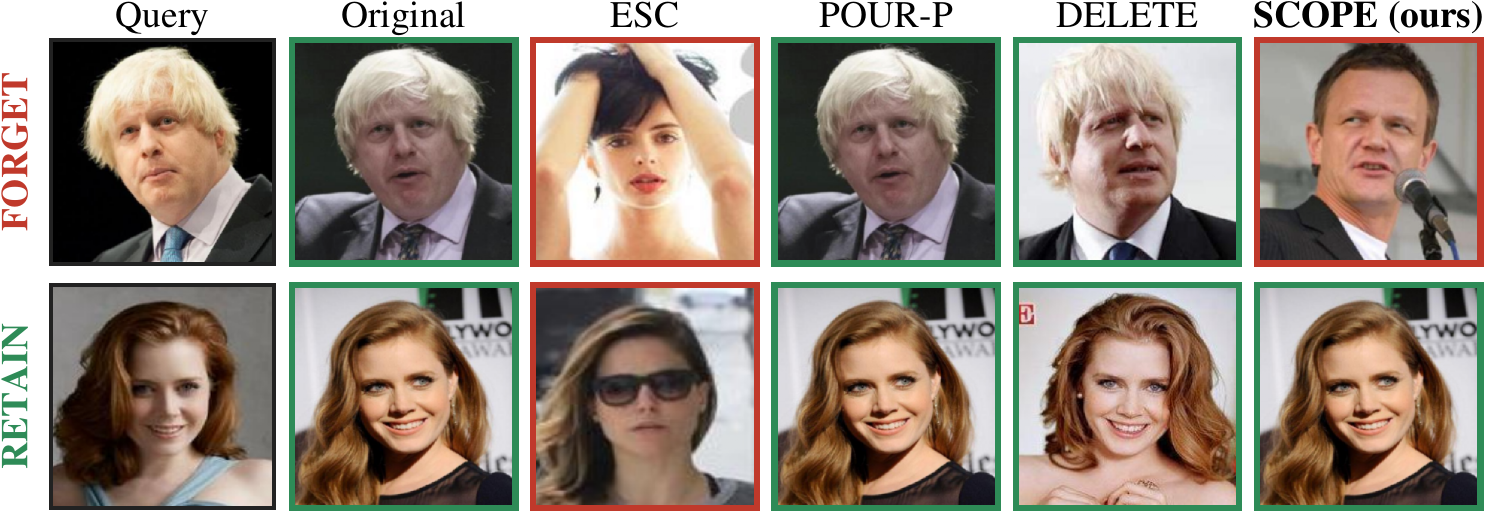}
    \caption{Rank-1 face retrieval~\citep{facenet} after deletion (VGGFace2-100, Swin-T, $K{=}20$; green/red borders mark same/different identity).}
    \label{fig:retrieval}
\end{figure}

\begin{table}[t]
\centering
\setlength{\tabcolsep}{2pt}
\renewcommand{\arraystretch}{1.05}
\begingroup
\tablefontsize
\begin{tabular}{@{}l c c c@{}}
\toprule
Variant & $D_{rt}^{\mathrm{KR}}\,\uparrow$ & $D_{ft}^{\mathrm{KR}}\,\downarrow$ & $HM_t^{\mathrm{KR}}\,\uparrow$ \\
\midrule
\multicolumn{4}{@{}l}{\emph{Gate scale (erased basis fixed)}} \\
\quad No gate (global) & \tabpm{80.9}{0.2} & \bestpm{0.0}{0.0} & \tabpm{89.4}{0.1} \\
\quad Hard limit ($\tau{\to}\infty$) & \secondpm{92.3}{0.5} & \bestpm{0.0}{0.0} & \bestpm{96.0}{0.2} \\
\quad Deployed (\textbf{SCOPE}) & \bestpm{92.4}{0.5} & \bestpm{0.0}{0.0} & \bestpm{96.0}{0.2} \\
\midrule
\multicolumn{4}{@{}l}{\emph{Erased basis (soft gate fixed)}} \\
\quad Random rank-$q$ basis & \tabpm{92.1}{0.3} & \tabpm{89.4}{3.5} & \tabpm{18.9}{5.7} \\
\quad $\hUF$ only & \secondpm{92.3}{0.5} & \tabpm{0.9}{0.6} & \tabpm{95.6}{0.5} \\
\quad $+$ variance & \secondpm{92.3}{0.4} & \secondpm{0.1}{0.1} & \secondpm{95.9}{0.2} \\
\quad $+\,\omega$ (\textbf{SCOPE}) & \bestpm{92.4}{0.5} & \bestpm{0.0}{0.0} & \bestpm{96.0}{0.2} \\
\bottomrule
\end{tabular}
\endgroup
\caption{Ablation on VCTK (Swin-T) at $K{=}10$ (mean$\pm$std over trials; best \textbf{bold}, second \underline{underlined} per column across all seven variants).}
\label{tab:ablation}
\end{table}

\subsection{Ablation Study}
\label{sec:exp-ablation}

SCOPE's two design choices act on separate axes (Table~\ref{tab:ablation}, the most entangled pair at $K{=}10$). The gate governs retention. Removing it collapses SCOPE onto the global projection, which still deletes but pays the frontier retain cost, whereas the hard and finite-$\tau$ gates are interchangeable up to trial noise. SCOPE's retention also holds under distribution shift (in the appendix). The basis governs deletion. With the same gate, a random rank-matched subspace keeps retention yet leaves the forget classes re-extractable ($D_{ft}^{\mathrm{KR}}$ $89.4$ against SCOPE's $0.0$), so the erased basis must carry the forget-discriminant structure. Cost-aware ranking helps, since variance ordering leaves more re-extractable forget accuracy than $\omega$ at no retention gain, and building the basis in explicitly whitened coordinates instead changes the score by at most $0.8$ points across ten pairs whose retain second moment's distance from the identity $\frac{\lVert\Sig_{\calR}{-}\I\rVert_F}{\sqrt d}$ spans $1.42$ to $6.35$ (in the appendix). Even the trivial maps the same gate admits, collapsing gated features to zero or a constant, leave the forget classes re-extractable at the $\tfrac1K$ floor because the collapsed point stays readable. SCOPE's erased basis instead drives re-extraction to zero, the gap widest at $K{=}1$, so the basis, not the shared gate, carries the erasure (in the appendix). Both the gate's retention ordering and the basis's deletion ordering persist across all fifty settings, with SCOPE within $0.1$ points of the best $HM_t^{\mathrm{KR}}$ in every column (in the appendix). Neither choice is tuned. Sweeping $\tau$ over $\{1,2,4,8,16,\infty\}$, the erased rank from $\hat s_{\calF}$ to twice the schedule $q_{\mathrm{sched}}(K)$, and the candidate pool $r_{\mathrm{pool}}$ over $\{32,64,128,256\}$ improves on the deployed configuration by at most $0.16$ points at every multi-class $K$, while a cumulative-forget-variance rule picks a median of $3.7$ to $6.2$ times that rank without improving it (in the appendix). SCOPE therefore deploys a single fixed configuration without hyperparameter search (in the appendix).

\section{Conclusion}
\label{sec:conclusion}
We show that source-free class deletion meets a fundamental entanglement frontier, a provable retain cost binding every fixed deleting linear map, and that conditioning the erasure per input escapes it by design. SCOPE turns this principle into a closed-form unlearner that deletes as deeply as retraining even under nearly all the stronger nonlinear probes. More broadly, we establish representational geometry as a general, predictive principle for selective, source-free forgetting.

\bibliography{refs.bib}

\end{document}